\documentclass[letterpaper, 10 pt, conference]{ieeeconf}  

\IEEEoverridecommandlockouts                              

\usepackage[]{graphicx}
\graphicspath{ {./figs/} }
\usepackage[outdir=./]{epstopdf}
\usepackage{multirow}
\usepackage{url}
\usepackage{amsmath}
\usepackage{amsfonts}
\usepackage{algorithm,algorithmicx}
\usepackage{algpseudocode}
\usepackage{verbatim}
\usepackage{cite}

\title{\LARGE \bf
Challenges in Partially-Automated Roadway Feature Mapping Using Mobile Laser Scanning and Vehicle Trajectory Data
}

\author{Mohammad Billah$^{\dagger}$, Farzana S. Rahman$^{\dagger}$, Arash Maskooki, Michael Todd, Matthew Barth and Jay A. Farrell
\thanks{*This work was  supported by a grant from the Cooperative Transportation Systems Pooled Fund Study subaward number GS11191-147223.}
\thanks{$^{\dagger}$The first two authors should be considered as joint first authors.}
\thanks{Billah and Rahman are with the department of Electrical and Computer Engineering,
        University of California, Riverside, 92521, U.S.A.
        {\tt\small \{mbillah, frimi\}@ece.ucr.edu}}%
\thanks{Todd is with the Center for Environmental Research \& Technology,
    	University of California, Riverside, 92521, U.S.A.
    	{\tt\small mike@cert.ucr.edu}}%
\thanks{Barth and Farrell are faculty of Electrical and Computer Engineering,
		University of California, Riverside, 92521, U.S.A.
        {\tt\small \{barth, farrell\}@ece.ucr.edu}}%
}

\begin{document}

\maketitle
\thispagestyle{empty}
\pagestyle{empty}


\begin{abstract}
Connected vehicle and driver's assistance applications are greatly facilitated by Enhanced Digital Maps (EDMs) that represent roadway features (e.g., lane edges or centerlines, stop bars).
Due to the large number of signalized intersections and  miles of roadway, manual development of EDMs on a global basis is not feasible. 
Mobile Terrestrial Laser Scanning (MTLS) is the preferred data acquisition method to provide data for automated EDM development.
Such systems provide an MTLS trajectory and a point cloud for the roadway environment.
The challenge is to automatically convert these data into an EDM.
This article presents a new processing and feature extraction method, experimental demonstration providing SAE-J2735 map messages for eleven example intersections, and a discussion of the results that points out remaining challenges and suggests directions for future research.

\end{abstract}

\section{Introduction}
\label{INTRO}
%
%
%
%


Recent technology advances have enabled consideration of cooperative applications between intelligent vehicles and the roadway infrastructure. 
Such cooperative applications are facilitated by roadway feature maps.
An accurate map of the signalized roadway enables: 
guidance and instructions to the user, 
optimization of intersection signaling based on nearby vehicles and their intentions,
warnings about specific road obstacles or hazards, and 
definitions of the territorial limits for traffic flows  thus ensuring a safe and comfortable driving environment.

There are over 300,000 signalized intersections and six million miles of roadway in the USA alone \cite{gordon2010traffic}.
For global commercialization, such maps would be required for all countries interested in participating; therefore automation of the roadway feature mapping is critically important.
Various sensors (e.g., radar, camera, and LiDAR) have been considered for feature sensors. 
Each has strengths and weaknesses, for example: weather condition, time of the day and complex shadowing from different objects \cite{ mccall2006video, haala2008mobile}. 
Due to advances in LiDAR technology, there has been increasing interest in Mobile Terrestrial Laser Scanning (MTLS).
MTLS has been adopted by both commercial and academic sectors as it captures highly precise and dense point clouds in relatively short periods of time \cite{haala2008mobile,tao2007advances, ussyshkin2009mobile, darnel2012using}.

This paper presents a complete point cloud to  roadway feature  extraction and mapping method.
The method is demonstrated using data from the California ITS testbed in Palo Alto which contains 11 intersections.
As is specified for connected vehicle applications, the intersection maps are output in SAE-J2735 \cite{dsrc2009dedicated} map message format.

%


Section \ref{RELEVANTWORK} reviews the related literature.
Section \ref{PROCESS} provides an overview of the process; describes the testbed, data, and data acquisition; and the georectification process. 
Section \ref{EXTRACTION}, which contians the novel ideas of the paper, states point cloud feature extraction problem and describes our approach.
Section \ref{RESULT} discusses the experimental results. 
Section \ref{CONCLUSION} concludes the study and makes suggestions for future work.

\section{Related Literature}
\label{RELEVANTWORK}

Several researchers have previously considered the use of LiDAR point cloud data for roadway feature mapping. 

Guan et. al. \cite{guan2015automated} extracts road curbs using the principle point selection method based on height values in short road slice segments. 
All the points inside road curbs are considered as the road surface. 
Then, an Inverse Distance Weighted (IDW) method is employed to generate a georectified raster image. 
Finally, high intensity road markings are detected in the raster image using Otsu's thresholding method.  
Kumar et. al. \cite{kumar2013automated} generates a 2D image from the 3D point cloud using height, intensity and pulse width attributes and detects road edges by implementing a parametric active contour model on the image. 
Then,  a range dependent thresholding function is employed to distinguish road markings from the image. 
Wang et. al. \cite{wang2015road} detects road curb lines using a saliency map computed from the 3D point cloud. Saliency is quantified by projecting the distance from each point’s normal vector to the point cloud’s dominant normal vector. 
Li et. al. \cite{wang2012automatic} uses statistical hypothesis testing on the altitude of each point cloud element to detect the boundary of the road. 
Yang et. al. \cite{yang2012automated} detects road curbs from a feature image computed from the point cloud. 
The feature image is generated by dividing the point cloud into small cells and interpolating each cell's gray value using IDW method. 
Then, the largest inter-cluster variance is used as the threshold to extract road surface. 
Finally, a Hough transform \cite{duda1972use} is performed to detect curbs from image line segments. 
Yuan et al. \cite{yuan2008road} extracts the road surface from MTLS LiDAR data using a fuzzy clustering method and fitting straight lines to the linearly clustered data using slope information. 
Lam et al. \cite{lam2010urban} extracts road points from MTLS LiDAR data by fitting RANdom SAmple Consensus (RANSAC) \cite{fischler1981random} planes to small sections and using a Kalman filtering to interconnect these fitted planes. 
Jaakkola et al. \cite{jaakkola2008retrieval} detects the road surface by filtering  the gradient image of the height attributes. 
Then the variance of the measured intensity value along each scan profile is reduced based on the fading of intensity. The points that are at a long distance from the laser scanner and large angle of incidence is removed. 
Then, road markings are detected by applying a threshold and morphological filtering method.  
Finally, the markings are classified according to area, area of the bounding box, and orientation. 
Smadja et al. \cite{smadja2010road} extracts roads from LiDAR data by using RANSAC along with slope information. 
Then road markings are extracted by simple thresholding. 
McElhinney et al. \cite{mcelhinney2010initial} presents an algorithm for extracting road edges from MTLS LiDAR data where a set of lines are fit to the road cross-sections based on the trajectory data and then LiDAR points within the vicinity of the lines are determined.
Yu et al.\cite{yu2015learning} presents an algorithm for detecting and classifying road markings directly from 3D point cloud data. 
First, a multi-segment thresholding and spatial density filtering is used in the reduced road surface points. 
Then different types of road markings are categorized through an integration of Euclidean distance based clustering, normalized cut segmentation, classification using trajectory and curb-lines, deep learning, and principal component analysis.

\section{Process Overview}
\label{PROCESS}

The MTLS process includes three major steps: data collection,  trajectory estimation and georectification, and  feature extraction and mapping.    
A rigid platform containing a suite of sensors (e.g. an inertial measurement unit (IMU), a global navigation satellite system receiver (GNSS), and a LiDAR) is moved through an environment for {\em data collection}.
The IMU and GNSS data are combined to provide the position and attitude (i.e. the {\em trajectory}) of the sensor platform to centimeter-level accuracy, which is needed to {\em georectify} the LiDAR point cloud data. 
{\em Feature extraction} is performed on the georectified point cloud.

Data collection, trajectory estimation and georectification are briefly reviewed in this section.
The main novel contributions of this paper are a feature extraction process and its experimental demonstrations.
Each of these is presented subsequently in its own section.


\subsection{Data Collection}

To facilitate reliable automated feature extraction, a goal is to ensure a high-density of LiDAR reflections from roadway relevant features. 
Therefore, the mapping vehicle should traverse  each intersection entry and exit point several times. 
This results in feature reflections from various orientations and distances and decreases the likelihood that a feature is not detected due to occlusion by other vehicles on the roadway. 

The point cloud dataset used in this paper was collected by the Advanced Highway Maintenance \& Construction Technology Research Center (AHMCT) research center at UC Davis in collaboration with Caltrans.
The data were collected at California ITS testbed in Palo Alto, which contains $11$ intersections along El Camino Real.
The data collection process included occasional control points (known locations) marked by highly reflective materials that could be used both for instrument calibration and confirmation of the accuracy of mapped features.

\subsection{Trajectory Estimation and Georectification}

The data acquisition process with the MTLS provides IMU, GNSS, and LiDAR point cloud data.
The IMU and GNSS data are combined in an optimal Bayesian smoothing process \cite{vu2013centimeter} to estimate the sensor platform position and attitude trajectory at the high (i.e., 200 Hz) sampling rate of the IMU.
This trajectory is required for the point cloud georectification process.
The trajectory estimation accuracy determines to a large extent the accuracy of the resulting map.

A LiDAR provides the point cloud with positions measured in the time-varying LiDAR frame  $\{L\}$ \cite{ellum}.
Each element of the point cloud is a 4-tuple $(x_i^l, y_i^l, z_i^l, I_i)$, 
where ${}^L P_i = (x_i^l, y_i^l, z_i^l)$ contains the coordinates of the $i$-th LiDAR reflection along the LiDAR local axes, and $I_i$ is the intensity of the $i$-th reflected point.
Each intensity $I_i$ depends on the angle-of-incidence, distance, roughness, and the reflectance of the point where the laser hit \cite{takashi}. 

The georectification process converts each point ${}^L P_i$ from LiDAR frame to ${}^W P_i$ world frame $\{W\}$.
The result is referred to as the georectified point cloud, using the equation:
\begin{equation}\label{eq:georec}
{^W}\!P_i = {^W}\!P_B + {^W_B}\!R \Big({}^B\!{P_L} + {^B_L}\!R ^L\!{P_i} \Big) 
\end{equation}
The offset ${}^B\!{P_L}$ and rotation ${^B_L}\!R$ from LiDAR to body $\{B\}$ frame are constant calibration parameters determined when the platform is constructed.
The rotation ${^W_B}\!R$ from body to world frame and the position ${^W}\!P_B $ of the body frame relative to world frame are the outputs of the platform trajectory estimation process.

\section{Feature Extraction}
\label{EXTRACTION}
After calibration and georectification of the  point cloud, roadway features (e.g., stop bars, lane edges etc.) are extracted. 
The problem statement and our solution approach are described in the following sections.

\subsection{Problem Statement}
\label{PROBLEM}

The problem is to find a set of features $F$ within a point cloud $P = \{(X_j, Y_j, Z_j, I_j) \, \mid \, j = 1, 2, \cdots, N \}$ that was acquired along a vehicle trajectory $T$.
The point cloud contains the georectified position $(X_j, Y_j, Z_j)$ and intensity $I_j$ of $N$ LiDAR returns. 
Typically $N$ is very large, in the billions. 
The set of features $F$ includes such items as lane centerline nodes, lane widths, lane direction, and stop bar locations.

For our discussion, this process will be divided into the following subproblems for the $i$-th intersection:
\begin{enumerate}
	\item Extract the intersection point cloud $P_i$ and trajectory $T_i$.
	\item Extract a road surface point cloud $S_i$.
	\item Convert  $S_i$ to an set of images.
	\item Extract features using image processing.
	\item Output a SAE J2735 map message.
\end{enumerate}

\subsection{Solution Approach}
\label{SOLUTION}

\noindent
The solution to each subproblem is defined below.
The data acquisition and geo-rectification processes provide  the point cloud $P$ and trajectory $T$. 
The point cloud $P$ contains large amounts of irrelevant points - from vehicles, sidewalks, vegetation, etc. - along with the desired road feature points.
In addition, it is too large to work with in its entirety.
The first few steps focus on finding portions of the point cloud that are relevant to the road surface near an intersection.

{\bf 1) Intersection Point Extraction. }
The first step is to subdivide $P$ and $T$ into subsets relevant to each intersection. 
For each intersection, the sets $P_i$ and $T_i$ are defined as subset of $P$ and $T$ that  retain only the data that is within radius are $R$ from the center of the $i$-th intersection. 
The intersection center is assumed to be given. 
Once $P_i$ and $T_i$ are defined for all intersections then, the original data sets $P$ and $T$ can be decomposed as
\begin{equation}
P = \bigcup_{i = 1}^{11} P_i + Q   \textnormal{ and }  T = \bigcup_{i = 1}^{11} T_i + U
\end{equation}
where $Q$ and $U$ contain point cloud and trajectory data not relevant to the intersections, which will be ignored in subsequent processing.

\begin{figure}[bt]
	\begin{center}
	\begin{tabular}{cc}
		\includegraphics[height=0.75\linewidth]{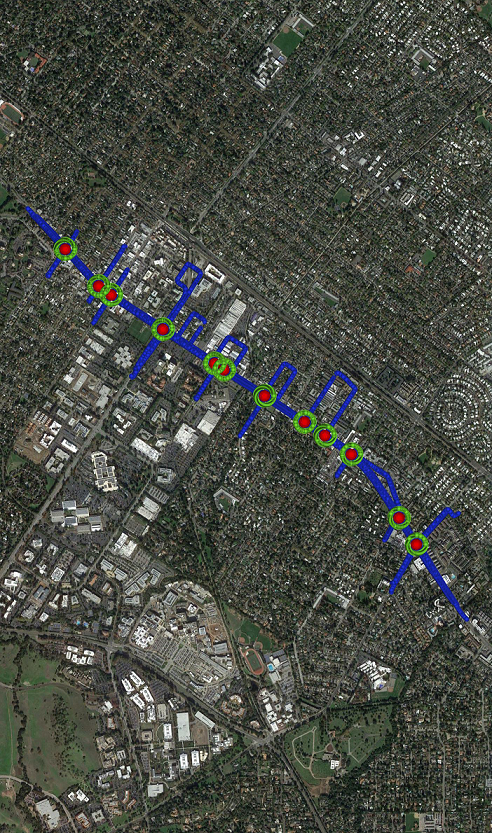} &
		\includegraphics[height=0.75\linewidth]{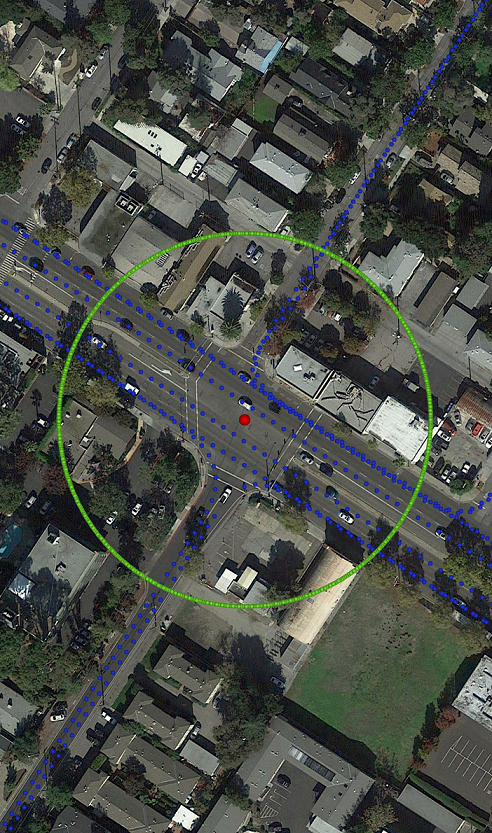} \\
		(a) & (b)
	\end{tabular}
	\end{center} 
	\vspace{-4mm}
	\caption{(a) Testbed trajectory including $11$  intersections. 
		(b) Trajectory $T_6$ for intersection $6$.} 
	 \label{fig:preproc}
\end{figure}

Fig.~\ref{fig:preproc}(a) shows the trajectory $T$ from the California ITS testbed. 
The eleven intersection centers are marked with red dots and green circles of radius $R = 60m$. 
Fig.~\ref{fig:preproc}(b) shows a magnified portion of $T_6$ in the vicinity of the 6-th intersection. 

{\bf 2) Road Surface Extraction.}
The inputs for this processing step are the sets $P_i$ and $T_i$. 
The goal of this step is to construct a reduced point cloud that only contains those points in $P_i$ that are on the roadway surface; thereby removing points reflected from extraneous objects (e.g., buildings, vehicles, signs).
This step contains three sub-steps: Road partitioning, road edge detection and road surface point extraction.

\begin{figure*}
	\begin{center}
		\begin{tabular}{ccc}
			\includegraphics[height=.23\linewidth]{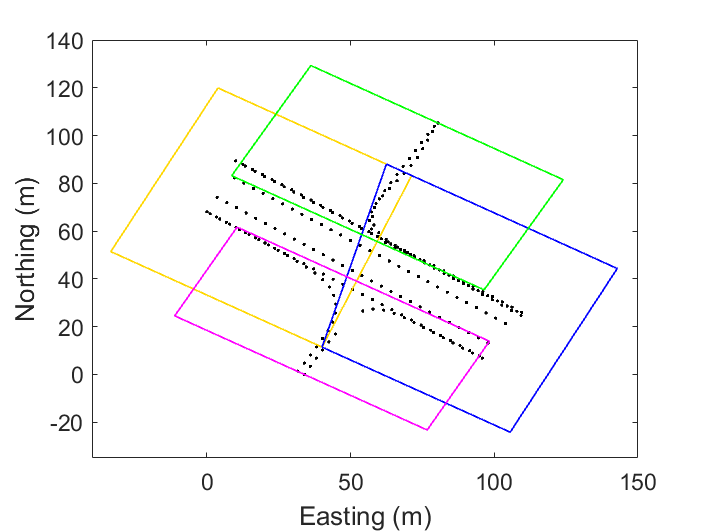} & \includegraphics[height=.23\linewidth]{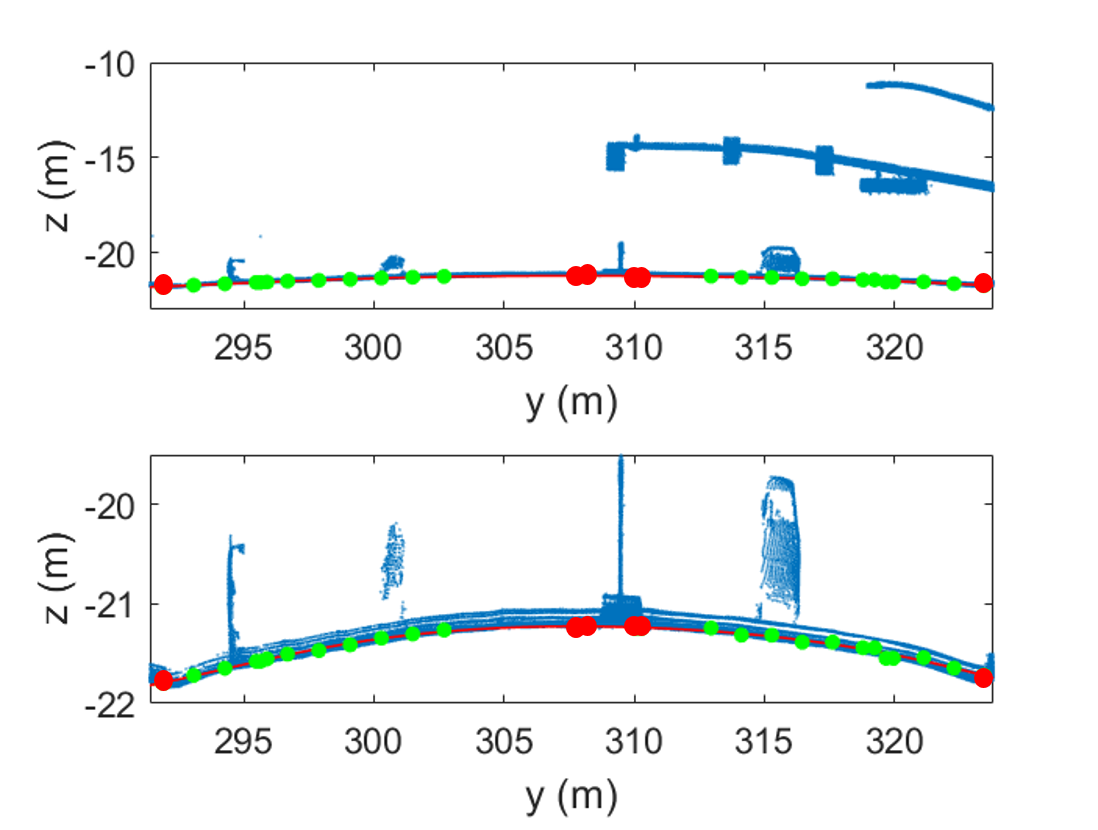} & \includegraphics[height=.23\linewidth]{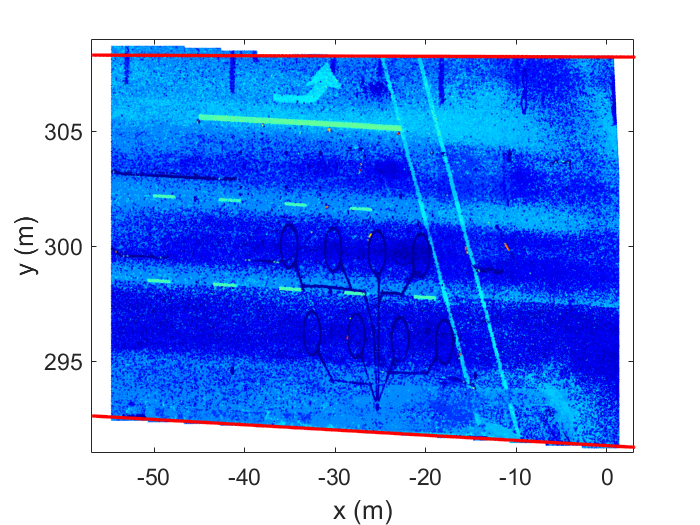}\\
			(a) & (b) & (c)
		\end{tabular}
	\end{center}
	\vspace{-4mm}
	\caption{(a) Road partitioning of an intersection. (b) Road surface points extraction from small cross-section of a partitioned road. (c) Extracted road surface and road edge points.  }
	\label{fig:surface}  
\end{figure*}

{\em 2a) Partition $P_i$ and $T_i$ into smaller road sections. } 
Initially, the only information about the location of the road is the trajectory  $T_i$ and the center location $C_i$.
Using this information, the region near the intersection can be partitioned into possibly overlapping polygons, such that each polygon contains the region of one roadway approaching the intersection.
A T-shaped intersection will yield $W=3$ rectangles and an X-shaped intersection will yield $W=4$ rectangles. 
The portion of $P_i$ within each polygon is denoted $P_i^w$.
Note that 
\begin{equation}
	P_i = \bigcup_{w=1}^{W}P_i^w + G_i
\end{equation}
where the points in $G_i$ are not along any of the approaches and will be discarded. 
Fig. \ref{fig:surface}(a)  shows the trajectory points for an intersection and the $4$ polygons defining the road sections.

{\em 2b) Process  $P_i^w$ to extract the road and median edges.} 
The idea of this step is that within a small enough region the road surface is a smooth 2D manifold.

Along the direction of motion of $T$ a one meter long rectangular swath of $P_i^w$ is extracted.
This point cloud swath $\Omega$ is much wider than the road.
Points in $\Omega$ may be reflected from the road surface or items off that surface.
Fig.  \ref{fig:surface}(b) shows an example portion of $\Omega$ where the view is along the $x$-axis, which points in the direction of motion.
The $z$-axis is vertical and the $y$-axis is horizontal and perpendicular to the direction of motion. 	

Within $\Omega$, the road surface is assumed to be two-dimensional.
Moving along the $y$-axis in five centimeter wide bins, the mode of the $z$ values is computed.
On the road surface, the mode of $z$  is a continuous function of $y$. 
At the road and median edges this function changes abruptly. 
These abrupt changes are detected by monitoring both the derivative and inflection points of the function. 
After extracting the road and meridian edges for a given $\Omega$, the rectangle is moved one meter along the $x$-axis and the processes is repeated.
This repetition produces points along the road and meridian edges as a function of $x$. 

After the above process completes for intersection section $w$, each set of road and median edge points is processed to remove outliers and insert estimates for missing data items. 
Then a piecewise line fitting approach is used to fit  road and meridian edge curves. 
After the road edge extraction, points outside the road edges are discarded to define the reduced point cloud $\bar{P}_i^w$ which along with the road and median curves are passed to the surface extraction process.

In Fig. \ref{fig:surface}(b) the red dots mark the detected road and meridian edges.
Note that at this point, even after discarding the points outside the road edges, the point cloud $\bar{P}_i^w$ still may contain points above the road surface.

{\em 2c) Extract the road surface point cloud for each road segment. } 
The objective of this  step is to extract from $\bar{P}_i^w$ a subset $S_i^w$ containing only  points on the road surface.

The platform trajectory points are first projected down onto the road surface using the known platform calibration parameters.
For each rectangular point cloud swath $\Omega$  defined in the previous step, some of these projected trajectory points will lie within it. 
These points are the initial seeds and are shown as green dots in Fig. \ref{fig:surface}(b). 
Additional road surface points are estimated between these projected points using the mode of the $z$ distribution as a function of $y$. 
Finally the road and edge points are included.
A 6-th order polynomial is fit to this set of points. 
Point cloud elements within $20 cm$ of this curve-fit are extracted as the road surface points.
This process repeats as $\Omega$ is slid along the trajectory.
The union of the road surface points over this sliding $\Omega$ regions defines the road surface point cloud denoted by $S_i^w$.

An example surface with the road edges is shown as a top-down view in Fig. \ref{fig:surface}(c).

{\bf 3) Map 3D points to 2D image. }
Many image processing tools are available to extract features from images. 
The purpose of this step is to convert each road surface point cloud $S_i^w$ into one or two raster images - one for each ingress or egress roadway section separated by a median.

\begin{figure}
	\begin{center}
		\begin{tabular}{c}
			\includegraphics[height=.38\linewidth]{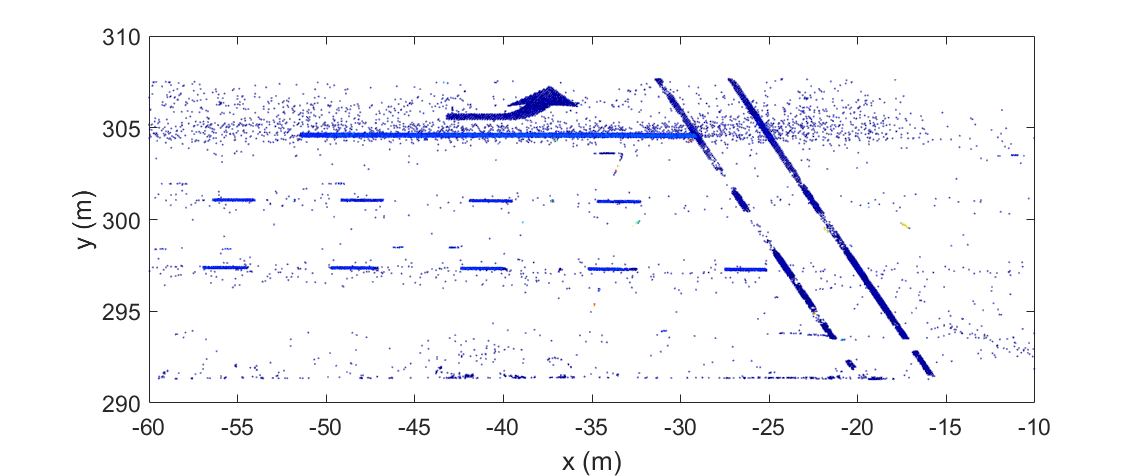}\\
			(a)\\
			\includegraphics[height=.31\linewidth]{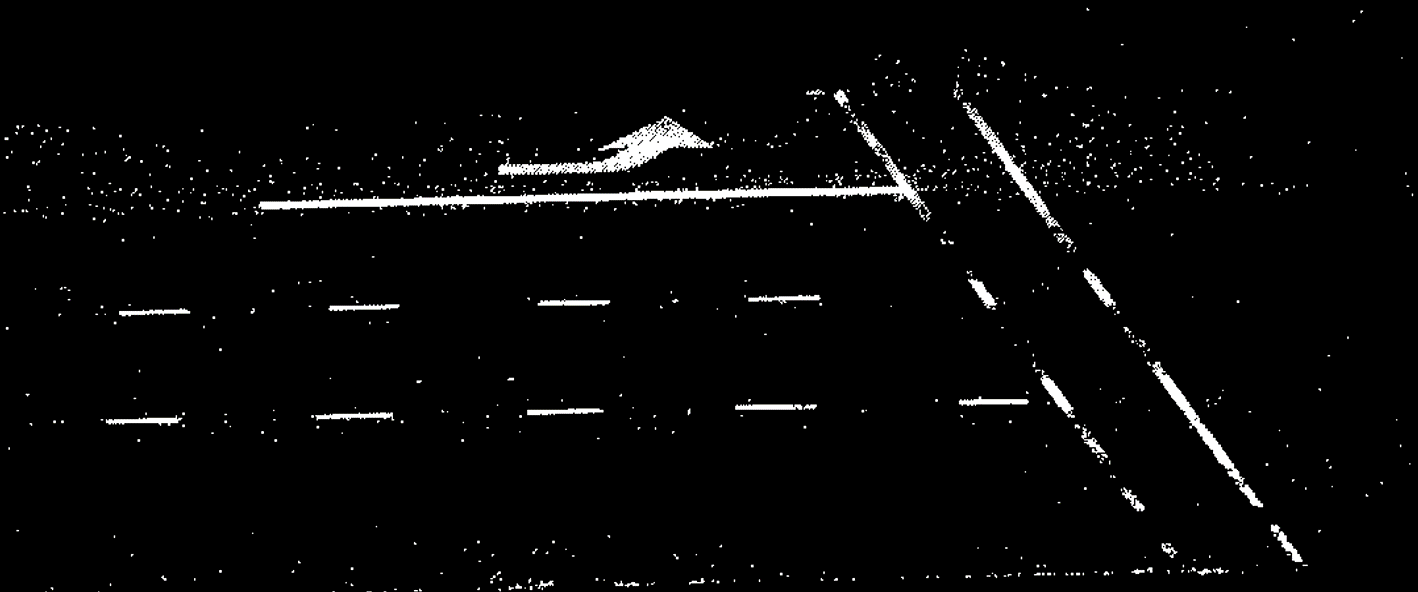} \\
			(b)
		\end{tabular}
	\end{center}
	\vspace{-4mm}
	\caption{(a) Point cloud $\bar{S}_i^{wk}$ after intensity thresholding. (b) Raster generated from thresholded point cloud.}
	\label{fig:3dto2d}  
\end{figure}

First, for roadway segments where a median was detected, the median curve is used to divide $S_i^w$ into two separate branches $S_i^{wk}$ for $k=1, 2$. 
If a road segment does not have a median, then the segment is treated as one branch with $k=1$.
Second, for each road surface segment branch $S_i^{wk}$, points with low reflectivity are removed from $S_i^{wk}$ to generate a new point cloud subset
\begin{equation}
\bar{S}_i^{wk} = \{ x \in S_i^{wk}  \, \mid \,  I(x) > \tau \}
\end{equation}
where $\tau$ is a user defined intensity threshold. 
The union of the intensity thresholded road surface segments generates an intensity thresholded point cloud $\bar{S}_i$ for the whole intersection.
A segment after intensity thresholding is shown in Fig. \ref{fig:3dto2d}(a). 

A 2D raster image $\bar{R}_i^{wk} $ is generated from each point cloud $\bar{S}_i^{wk} $. 
The pixel size of the raster image in meters is a trade-off between computational complexity and feature position estimation accuracy.
For the feature extraction approach described herein, each raster pixel is $3 cm \times 3 cm$. 
Therefore, each raster pixel maps to a $3 cm$ square cell in the XY plane. 
The value of the pixel for each cell depends on the elements of $\bar{S}_i^{wk}$ that map to the cell. 
When no elements of $\bar{S}_i^{wk}$ map to a cell, the pixel value is zero. 
Otherwise, the pixel value is the average value of the intensities of the elements of $\bar{S}_i^{wk} $ that map to the cell. 
The raster image corresponding to Fig. \ref{fig:3dto2d}(a) is shown in Fig. \ref{fig:3dto2d}(b).

\begin{figure*}
	\begin{center}
		\begin{tabular}{ccc}
			\includegraphics[height=.15\linewidth]{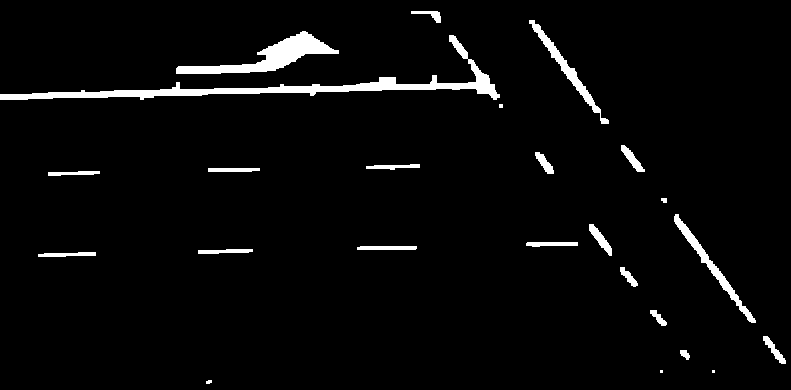} & \includegraphics[height=.15\linewidth]{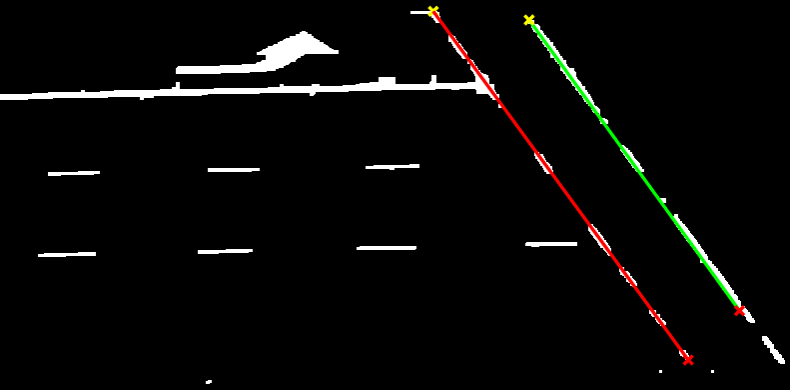} & \includegraphics[height=.15\linewidth]{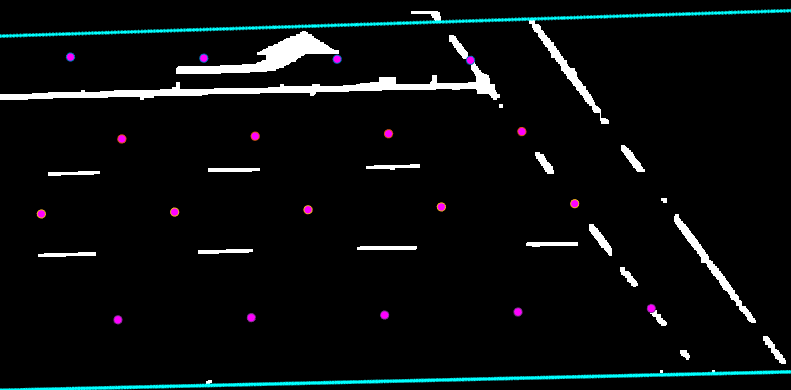}\\
			(a) & (b) & (c)\\
		\end{tabular}
	\end{center}
	\vspace{-4mm}
	\caption{(a) Raster image of an ingress road section (b) Stop bar (green solid line) (c) lane node points (magenta dots) and road edge line (cyan solid line) detection }
	\label{fig:featureExtraction}  
\end{figure*}

{\bf 4) Image-based Roadway Feature Extraction.}
This section describes: 
image processing to improve  feature detectability,
detection of stop bars,
detection of lane dividers,
definition of lane centerlines and lane widths.
The input images are $\bar{R}_i^{wk}$ for the various values of $i$, $w$, and $k$.

Noise and unwanted artifacts  degrade the performance of feature recognition; therefore, morphological filters (erosion and dilation) are first used to reduce the number of artifacts present. 
Lines are extracted using the Hough transform. 
Lines with sufficient length that are oriented approximately orthogonal to the trajectory direction are extracted as potential stop bars. 
More than one line may fit these criteria when a pedestrian crosswalk exists, then the line second farthest from the intersection is classified as stop-bar.

Lines parallel to the direction of traffic flow are candidates for lane dividers. 
The algorithm uses a decision tree to process the lines returned by the Hough transform and the meridian and road edges. 
First, depending on the width of $\bar{R}_i^{wk}$, the algorithm labels the branch as single or multiple lanes. 
Then the algorithm first checks for solid lines, as used to demarcate turning lanes. 
After finding solid lines, the remaining road width is computed to determine the maximum number of remaining lanes. 
Given this estimate of the maximum number of remaining lanes, the algorithm begins searching through the remaining lines. 
The process concludes either when the remaining maximum number of remaining lanes is less than one or when there are no remaining suitable lines to serve as lane dividers. 
In some cases, for cross-streets, the branch $\bar{R}_i^{wk}$ is too narrow to support multiple lanes and does not have lane markings. 
In this case, the lane edges are defined to be the road and or median edges. 
In some other cases, such as cross-streets, the branch has two lanes with no detectable markings, yet has width sufficient for two lanes. In this case, the algorithm defines the branch to have two lanes.

Once the lane edges are determined, the lane centerline is computed by putting nodes midway between the two lane edges. Nodes are separated by a distance of approximately $D = 6 m$ (200 pixels apart).  
The first node of the centerline is defined to be on the stop-bar. 
Fig. \ref{fig:featureExtraction}c shows the output of the feature extraction algorithm. 
The input image $\bar{R}_i^{wk}$ is the background. 
The input road edges are shown as green solid lines. 
The extracted stop bar is shown as a solid red line and extracted centerline nodes as magenta dots.

{\bf 5) SAE J2735 Intersection Feature Map Output. }
The SAE J2735 map message uses the lane centerline nodes, lane widths and stop bar locations. 
The previous step estimated these items in image coordinates.
After image coordinates are converted to UTM coordinates using the metadata saved during raster generation, the J2735 map message is generated.

\section{Results}
\label{RESULT}

The  MTLS used for data collection was a Trimble MX8, which includes two Riegl VQ450 laser scanners and an Applanix POS 520 platform positioning system. 
The MTLS platform was driven along each ingress and egress section of the main thoroughfare (El Camino Real) of the Palo Alto testbed multiple times and along each cross street at least once. 
The point cloud $P$ contained 2,560 million points, approximately 872 million are within the radius $R=60m$ of at least one of the intersection centers. The remaining 1688 million points were discarded.

The process described in the previous sections has been applied to extract J2735 map message for eleven intersections. The J2735 map message includes coordinates for the intersection center, the number of lanes and  a data structure for each lane including a lane identifier, a sequence of nodes defining the lane centerline, the lane width, a point indicating the start of the centerline at the stop bar and lane attributes such as whether the lane is for ingress or egress.

\begin{figure}
	\begin{center}
		\begin{tabular}{cc}
			\includegraphics[height=0.45\linewidth]{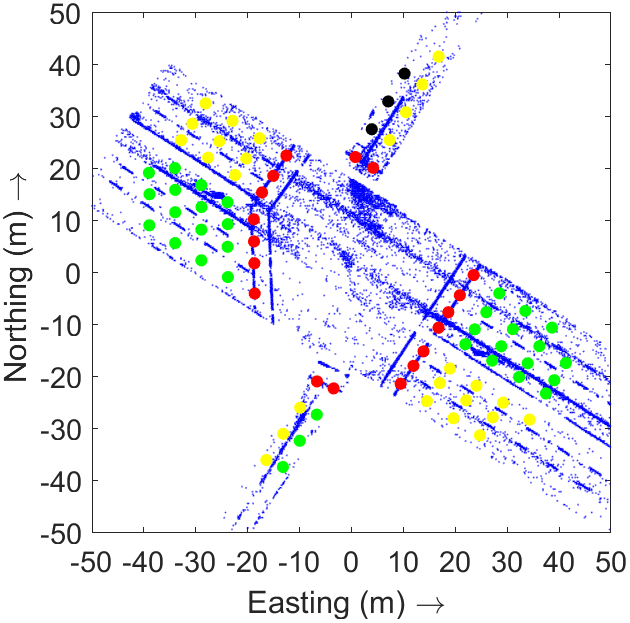} &
			\includegraphics[height=0.45\linewidth]{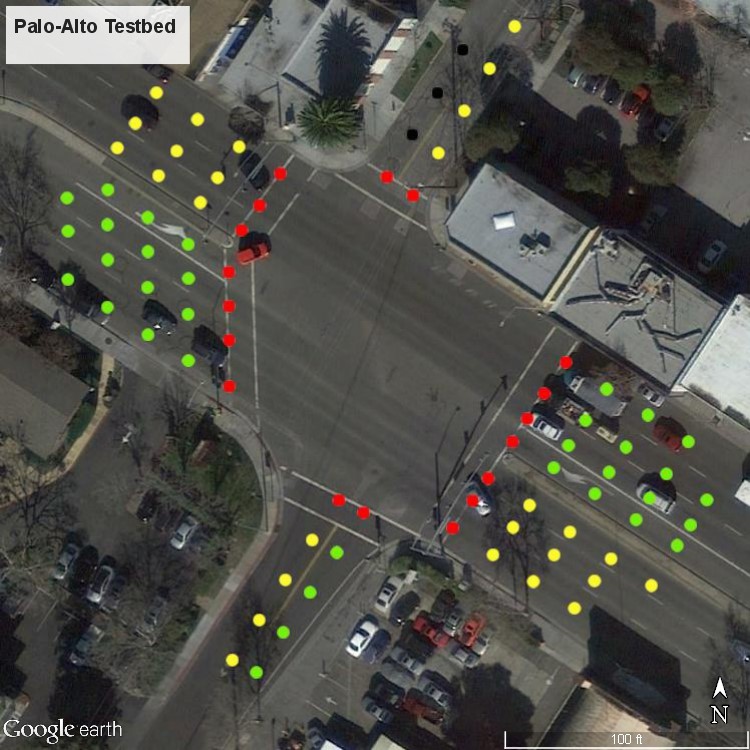} \\
			(a) & (b)
		\end{tabular}
	\end{center} 
	\vspace{-4mm}
	\caption{(a) J2735 node points superimposed on an intersection's point cloud (b) J2735 node points superimposed on Google Earth image} 
	\label{fig:result}
\end{figure}

Fig. \ref{fig:result}(a) shows the top-down view of the nodes in an extracted J2735 message superimposed on the intensity thresholded point cloud $\bar{S}_i$ from which it was extracted. 
According to the specification of the J2735, the first node of each lane starts at the position of the stop bar. 
This figure demonstrates the accuracy of the extracted J2735 map relative to $\bar{S}_i$, which is accurate to the centimeter level.
Fig. \ref{fig:result}(b) shows the same nodes  superimposed on a Google Earth image.
These figures show that the extracted J2735 message describes the complete intersection. 
The J2735 overlaid on the Google Earth image gives a visually interpretable result.\footnote{The Google Earth imagery is accurate to 2-5 meters \cite{mohammed2013positional, paredes2013horizontal, farah2014positional}. 
In some instances the entire J2735 overlay is offset relative to the Google Earth image while aligning with the point cloud $\bar{S}_i$.
Such cases demonstrated that the alternative approach of extracting intersection maps from Google Earth images could yield maps offset by a few meters. },
 The Google Earth  overlay is useful for detecting relative errors between portions of the J2735, such as the one  stop bar at the wrong location.

Standard T-shaped or X-shaped intersections process completely with correct outputs, when the road markings are not faded. 
In cases where the road markings are severely faded, the road has no markings, or the input image is very noisy the algorithm generates a warning and marks the segment for human processing. 
Another issue arises when only one stop bar is detected, in which case the algorithm proceeds with the detected stop bar and generates a warning for further human inspection. 
In case when no stop bar is detected the algorithm generates a warning so that the user can add a stop bar  manually.

\section{CONCLUSION}
\label{CONCLUSION}

This paper has presented an automated feature extraction approach with figures demonstrating its application.
The full set of detailed results for eleven intersections can be found at \cite{farrell2016best}.
On the continuum between manual and fully automated, the current algorithm is somewhere in the middle, human-assisted. When the algorithm detects an anomalous situation, it stops to alert the operator, this is considered good practice because the generated maps have implications for human safety.  

The analysis of the algorithm performance in  \cite{farrell2016best} shows that it works very well for standard intersections and when roadway markings are not too faded. 
Additional work is required to achieve functionality for non-standard intersections.  
Also, the algorithm includes parameters such as global intensity thresholds that currently require human adjustment.
Since the quality of the roadway markings and road surface vary locally, it would be beneficial to develop an automated method that adapts the intensity threshold locally, without human interaction.
Finally, the conversion of the point cloud to an image (rasterization) is an information losing step: position accuracy is lost due to the finite pixel size; intensity information is lost as many point cloud points are combined into one pixel value. 
This information loss would be avoided if methods are developed that work with the point cloud directly.





\section*{ACKNOWLEDGMENT}
The authors thank the personnel at the Advanced Highway Maintenance \& Construction Technology Research Center (AHMCT) for providing the 3D point cloud.

\bibliographystyle{IEEEtran}
\bibliography{bare_jrnl}

\begin{thebibliography}{10}
\providecommand{\url}[1]{#1}
\csname url@rmstyle\endcsname
\providecommand{\newblock}{\relax}
\providecommand{\bibinfo}[2]{#2}
\providecommand\BIBentrySTDinterwordspacing{\spaceskip=0pt\relax}
\providecommand\BIBentryALTinterwordstretchfactor{4}
\providecommand\BIBentryALTinterwordspacing{\spaceskip=\fontdimen2\font plus
\BIBentryALTinterwordstretchfactor\fontdimen3\font minus
  \fontdimen4\font\relax}
\providecommand\BIBforeignlanguage[2]{{%
\expandafter\ifx\csname l@#1\endcsname\relax
\typeout{** WARNING: IEEEtran.bst: No hyphenation pattern has been}%
\typeout{** loaded for the language `#1'. Using the pattern for}%
\typeout{** the default language instead.}%
\else
\language=\csname l@#1\endcsname
\fi
#2}}

\bibitem{gordon2010traffic}
R.~L. Gordon, \emph{Traffic signal retiming practices in the United
  States}.\hskip 1em plus 0.5em minus 0.4em\relax Transp. Research Board, 2010,
  vol. 409.

\bibitem{mccall2006video}
J.~C. McCall and M.~M. Trivedi, ``Video-based lane estimation and tracking for
  driver assistance: survey, system, and evaluation,'' \emph{IEEE T. on
  Intelli. Transp. Syst.}, vol.~7, no.~1, pp. 20--37, 2006.

\bibitem{haala2008mobile}
N.~Haala, M.~Peter, A.~Cefalu, and J.~Kremer, ``Mobile {LIDAR} mapping for
  urban data capture,'' \emph{14th Int. Conf. on Virtual Syst. and Multimedia},
  pp. 95--100, 2008.

\bibitem{tao2007advances}
C.~V. Tao and J.~Li, \emph{Advances in mobile mapping technology}.\hskip 1em
  plus 0.5em minus 0.4em\relax CRC Press, 2007, vol.~4.

\bibitem{ussyshkin2009mobile}
V.~Ussyshkin, ``Mobile laser scanning technology for surveying application:
  from data collection to end-products,'' \emph{FIG Working Week}, 2009.

\bibitem{darnel2012using}
C.~Darnel, ``Using {LIDAR} to solve industry challenges,'' \emph{Geo:
  Geoconnexion Int. Mag.}, vol.~11, no.~1, pp. 18--19, 2012.

\bibitem{dsrc2009dedicated}
\emph{Dedicated short range communications {(DSRC)} message set dictionary},
  SAE Int., {Tech. Rep.} J2735\_200911.

\bibitem{guan2015automated}
H.~Guan, J.~Li, Y.~Yu, M.~Chapman, and C.~Wang, ``Automated road information
  extraction from mobile laser scanning data,'' \emph{IEEE T. on Intelli.
  Transp. Syst.}, vol.~16, no.~1, pp. 194--205, 2015.

\bibitem{kumar2013automated}
P.~Kumar, C.~P. McElhinney, P.~Lewis, and T.~McCarthy, ``An automated algorithm
  for extracting road edges from terrestrial mobile {LIDAR} data,'' \emph{ISPRS
  J. of Photogramm. and Rem. Sens.}, vol.~85, pp. 44--55, 2013.

\bibitem{wang2015road}
H.~Wang, H.~Luo, C.~Wen, J.~Cheng, P.~Li, Y.~Chen, C.~Wang, and J.~Li, ``Road
  boundaries detection based on local normal saliency from mobile laser
  scanning data,'' \emph{IEEE Geoscience and Rem. Sens. Let.}, vol.~12, no.~10,
  pp. 2085--2089, 2015.

\bibitem{wang2012automatic}
H.~Wang, Z.~Cai, H.~Luo, C.~Wang, P.~Li, W.~Yang, S.~Ren, and J.~Li,
  ``Automatic road extraction from mobile laser scanning data,'' \emph{Int.
  Conf. on Comp. Vis. in Rem. Sens.}, pp. 136--139, 2012.

\bibitem{yang2012automated}
B.~Yang, Z.~Wei, Q.~Li, and J.~Li, ``Automated extraction of street-scene
  objects from mobile {LIDAR} point clouds,'' \emph{Int. J. of Rem. Sens.},
  vol.~33, no.~18, pp. 5839--5861, 2012.

\bibitem{duda1972use}
R.~O. Duda and P.~E. Hart, ``Use of the {H}ough transformation to detect lines
  and curves in pictures,'' \emph{Comm. of the ACM}, vol.~15, no.~1, pp.
  11--15, 1972.

\bibitem{yuan2008road}
X.~Yuan, C.~X. Zhao, Y.~F. Cai, H.~Zhang, and D.~B. Chen, ``Road-surface
  abstraction using {LADAR} sensing,'' \emph{10th Int. Conf. on Control,
  Automation, Robotics and Vis.}, pp. 1097--1102, 2008.

\bibitem{lam2010urban}
J.~Lam, K.~Kusevic, P.~Mrstik, R.~Harrap, and M.~Greenspan, ``Urban scene
  extraction from mobile ground based {LIDAR} data,'' \emph{3rd Int. Symp. on
  3{D} Data Proc., Visualization and Transp.}, pp. 1--8, 2010.

\bibitem{fischler1981random}
M.~A. Fischler and R.~C. Bolles, ``Random sample consensus: a paradigm for
  model fitting with applications to image analysis and automated
  cartography,'' \emph{Comm. of the ACM}, vol.~24, no.~6, pp. 381--395, 1981.

\bibitem{jaakkola2008retrieval}
A.~Jaakkola, J.~Hyypp{\"a}, H.~Hyypp{\"a}, and A.~Kukko, ``Retrieval algorithms
  for road surface modelling using laser-based mobile mapping,''
  \emph{Sensors}, vol.~8, no.~9, pp. 5238--5249, 2008.

\bibitem{smadja2010road}
L.~Smadja, J.~Ninot, and T.~Gavrilovic, ``Road extraction and environment
  interpretation from {LIDAR} sensors,'' \emph{IAPRS}, vol.~38, pp. 281--286,
  2010.

\bibitem{mcelhinney2010initial}
C.~McElhinney, P.~Kumar, C.~Cahalane, and T.~McCarthy, ``Initial results from
  european road safety inspection ({EURSI}) mobile mapping project,''
  \emph{ISPRS Commission V Technical Symp.}, vol. 2124, 2010.

\bibitem{yu2015learning}
Y.~Yu, J.~Li, H.~Guan, F.~Jia, and C.~Wang, ``Learning hierarchical features
  for automated extraction of road markings from 3{D} mobile {LiDAR} point
  clouds,'' \emph{IEEE J. of Sel. Topics in Applied Earth Obs. and Rem. Sens.},
  vol.~8, no.~2, pp. 709--726, 2015.

\bibitem{vu2013centimeter}
A.~Vu, J.~A. Farrell, and M.~Barth, ``Centimeter-accuracy smoothed vehicle
  trajectory estimation,'' \emph{IEEE Intelli. Transp. Syst. Mag.}, vol.~5,
  no.~4, pp. 121--135, 2013.

\bibitem{ellum}
C.~Ellum and N.~El-Sheimy, ``Land-based mobile mapping systems,''
  \emph{Photogramm. Engr. and Rem. Sens.}, vol.~68, no.~1, pp. 13--17, 2002.

\bibitem{takashi}
T.~Ogawa and K.~Takagi, ``Lane recognition using on-vehicle lidar,''
  \emph{Intelli. Vehicles Symp.}, pp. 540--545, 2006.

\bibitem{mohammed2013positional}
N.~Z. Mohammed, A.~Ghazi, and H.~E. Mustafa, ``Positional accuracy testing of
  {G}oogle {E}arth,'' \emph{Int. J. of Multidisciplinary Sciences and
  Engineering}, vol.~4, no.~6, pp. 6--9, 2013.

\bibitem{paredes2013horizontal}
C.~U. Paredes-Hern{\'a}ndez, W.~E. Salinas-Castillo, F.~Guevara-Cortina, and
  X.~Mart{\'\i}nez-Becerra, ``Horizontal positional accuracy of {G}oogle
  {E}arth's imagery over rural areas: a study case in {T}amaulipas, {M}exico,''
  \emph{Boletim de Ci{\^e}ncias Geod{\'e}sicas}, vol.~19, no.~4, pp. 588--601,
  2013.

\bibitem{farah2014positional}
A.~Farah and D.~Algarni, ``Positional accuracy assessment of {G}oogleearth in
  {R}iyadh,'' \emph{Artificial Satellites}, vol.~49, no.~2, pp. 101--106, 2014.

\bibitem{farrell2016best}
J.~A. Farrell, M.~Todd, and M.~Barth, ``Best practices for surveying and
  mapping roadways and intersections for connected vehicle applications,''
  Available at \url{http://escholarship.org/uc/item/4f88m75k}, 2016.

\end{thebibliography}

\end{document}